%% file: main.tex
\documentclass[10pt,journal,compsoc]{IEEEtran}

\ifCLASSOPTIONcompsoc
  \usepackage[nocompress]{cite}
\else
  \usepackage{cite}
\fi

\ifCLASSINFOpdf
\else
\fi

\usepackage{amsmath,amsfonts}
\usepackage{algorithmic}
\usepackage{array}
\usepackage[caption=false,font=normalsize,labelfont=sf,textfont=sf]{subfig}
\usepackage{textcomp}
\usepackage{stfloats}
\usepackage{url}
\usepackage{verbatim}
\usepackage{graphicx}
\usepackage{comment}
\usepackage{multirow}
\usepackage{makecell}
\usepackage{tabularx}
\def\BibTeX{{\rm B\kern-.05em{\sc i\kern-.025em b}\kern-.08em
    T\kern-.1667em\lower.7ex\hbox{E}\kern-.125emX}}
\usepackage{balance}
\begin{document}
\title{Quantum-Brain: Quantum-Inspired Neural Network Approach to Vision-Brain Understanding}
\author{
Hoang-Quan Nguyen$^1$, Xuan-Bac Nguyen$^1$, Hugh Churchill$^2$, \\
Arabinda Kumar Choudhary$^3$, Pawan Sinha$^4$, Samee U. Khan$^5$, Khoa Luu$^1$ \\
$^1$Department of Electrical Engineering and Computer Science, University of Arkansas, AR\\
$^2$Department of Physics, University of Arkansas, AR\\
$^3$Department of Radiology, SUNY Upstate Medical University, NY\\
$^4$Department of Brain and Cognitive Sciences, Massachusetts Institute
of Technology, MA\\
$^5$Department of Electrical and Computer Engineering, Kansas State
University, KS\\
\small{\texttt{\{hn016, xnguyen, hchurch, khoaluu\}@uark.edu}} \\
\small{\texttt{choudhaa@upstate.edu}} \quad \small{\texttt{psinha@mit.edu}} \quad \small{\texttt{sameekhan@ksu.edu}}
}

\markboth{Journal of \LaTeX\ Class Files,~Vol.~18, No.~9, September~2020}%
{How to Use the IEEEtran \LaTeX \ Templates}

\maketitle

\begin{figure*}[t]
    \centering
    % \captionsetup{type=figure}
    % \setlength{\abovecaptionskip}{-1pt}
    % \vspace{-4mm}
    \includegraphics[width=0.95\linewidth]{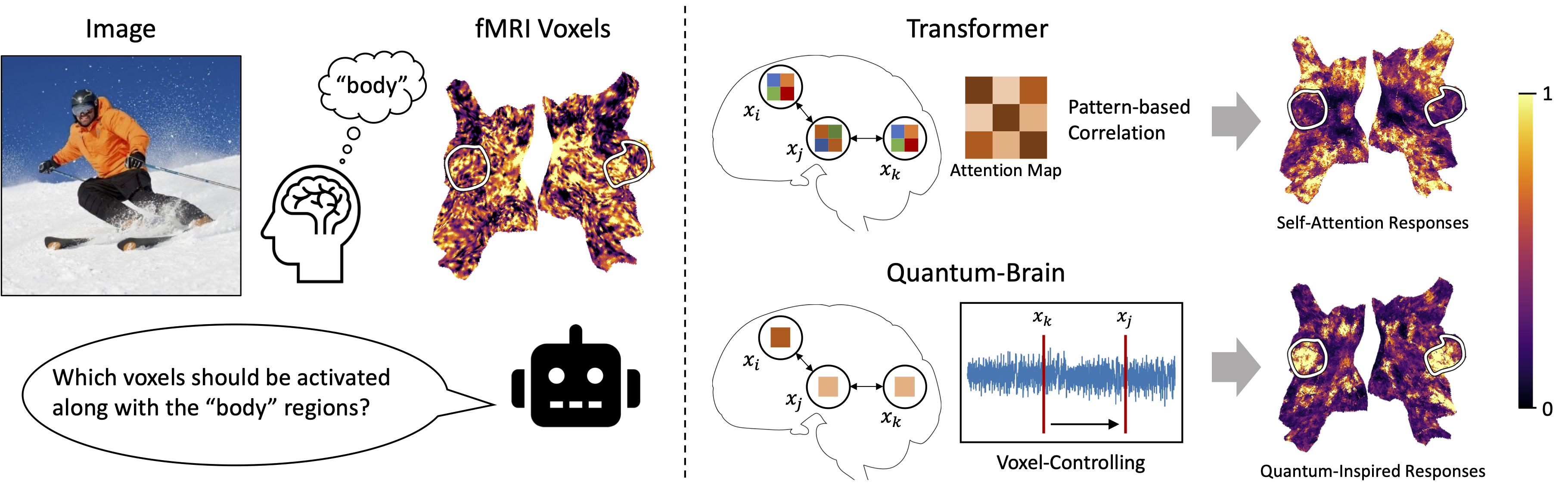} 
    % \vspace{-2mm}
    \caption{
    \textbf{Quantum-Brain for Vision-Brain Understanding.}
    Motivated by the connectivities in brain signals and entanglement in quantum theory, we propose Quantum-Brain, a quantum-inspired neural network to explore the connectivities in brain signals for vision-brain understanding.
    Apart from self-attention, we evaluate the entanglement between brain voxels to enhance the exploration of connectivity.}
\label{fig:intro}
% \vspace{-2mm}
\end{figure*}

\begin{abstract}
Vision-brain understanding aims to extract semantic information about brain signals from human perceptions.
Existing deep learning methods for vision-brain understanding are typically introduced within a traditional learning paradigm, lacking the ability to learn the connectivities between brain regions. Meanwhile, the quantum computing theory offers a new paradigm for designing deep learning models in classical systems. 
Motivated by the connectivities in brain signals and the entanglement properties in quantum computing, we propose a novel Quantum-Brain approach, a quantum-inspired neural network, to tackle the vision-brain understanding problem. 
To compute the connectivity between areas in brain signals, we introduce a new Quantum-Inspired Voxel-Controlling module to learn the impact of a brain voxel on others represented in the Hilbert space.
To effectively learn connectivity, a novel Phase-Shifting module is presented to calibrate the value of the brain signals.
Finally, we introduce a new Measurement-like Projection module to present the connectivity information from the Hilbert space into the feature space. The proposed approach can learn to find the connectivities between fMRI voxels and enhance the semantic information obtained from human perceptions.
Our experimental results on the Natural Scene Dataset benchmarks demonstrate the effectiveness of the proposed method, achieving Top-1 accuracies of $95.5\%$ and $95.3\%$ for image and brain retrieval tasks, respectively, and an Inception score of $95.6\%$ for the fMRI-to-image reconstruction task.
Our proposed quantum-inspired network introduces a novel paradigm for solving vision-brain problems using quantum computing theory.
\end{abstract}


\section{Introduction}

Human perception of the world is determined by external stimuli and individual experiments that create brain activities involving non-linear interactions among 86 billion neuronal cells, which are thus complex \cite{olshausen1996emergence,rolls1995sparseness}.
An effective method called functional Magnetic Resonance Imaging (fMRI) is typically used to measure brain activity via Blood Oxygenation Level Dependent (BOLD) signals, implicitly revealing correlations or connectivities between neurons in response to stimuli \cite{van2010exploring,zhou2010divergent}.
With the development of deep learning models, it is essential to learn the information contained in the measured fMRI signals, along with their corresponding visual data \cite{shen2019deep,beliy2019voxels}.

Vision-brain understanding aims to learn information from brain signals obtained from image stimuli.
It offers the exploration interest of cognition and perception that potentially contributes to brain-computer interface  \cite{du2022fmri} and beyond.
Brain-to-image, especially fMRI-to-image, is one of the fundamental tasks in understanding the brain and vision.
Generative models including GANs \cite{goodfellow2020generative} and diffusion models \cite{rombach2022high,xu2023versatile} enable to reconstruct more realistic images \cite{shen2019deep,ozcelik2022reconstruction,takagi2023high,scotti2024reconstructing}.
Moreover, cross-subject and multiple-subject brain decoding is taken into account \cite{wang2024mindbridge,scotti2024mindeye2,quan2024psychometry}.
Besides, image and brain retrievals are also essential to map the information of the brain signals and images into the same information space \cite{lin2022mind,ozcelik2023natural,scotti2024reconstructing}.
However, understanding the vision-brain connection is still confronted with significant challenges that hinder its broader-scale applications.

\noindent
\textbf{Limitations in Prior Methods:}
Prior methods focus on conditioned image generation models \cite{takagi2023high,sun2024contrast} that overlook information extraction from brain signals.
The other methods focus on the fMRI activity mapping to the embeddings of the images \cite{scotti2024reconstructing,chen2023seeing,quan2024psychometry,wang2024mindbridge}.
However, these methods make assumptions about the roles of brain regions that \textit{hinder the exploration of complex information and connectivity in brain signals}.
Moreover, since the connectivities in the fMRI signals are crucial to extracting the information, the multi-layer perceptron (MLP) backbone models in prior methods \cite{scotti2024reconstructing,scotti2024mindeye2} \textit{show the missing correlation between fMRI voxels in the brain activities}.
Besides, self-attention is one of the conventional approaches to computing the correlation between elements \cite{quan2024psychometry,chen2023seeing}.
However, in fMRI voxels, the connectivities are based on the functional information of each brain region. At the same time, self-attention learns to find patterns for correlation computing, which makes the model \textit{harder to learn and extract the connectivity}.
Fig. \ref{fig:intro} illustrates the difference between self-attention and the proposed approach.

\noindent
\textbf{Motivations from Quantum Machine Learning:}
Far apart from traditional computers, quantum computers are emerging machines that perform quantum algorithms \cite{arute2019quantum}. 
Quantum computing utilizes quantum theory to process data in quantum devices \cite{hidary2019quantum,nielsen2010quantum} with existing models including quantum circuits, quantum annealing, and adiabatic quantum computation. 
It has been proved that quantum computers outperform classical computers in solving specific problems \cite{grover1996fast,shor1999polynomial}. 
In the noisy intermediate-scale quantum (NISQ) era \cite{preskill2018quantum}. In contrast, quantum computers cannot perform complex quantum algorithms for practical applications, the quantum computing theory provides a new mathematical formalism for computational tasks. 
Hence, quantum machine learning is introduced as a new learning paradigm utilizing quantum computation to enhance classical machine learning models \cite{schuld2015introduction,biamonte2017quantum}.
Moreover, quantum theory has been applied to classical algorithms and deep learning models, which are expected to improve computational performance and quality \cite{gkoumas2021quantum,tang2022image}.
Furthermore, prior clinical and theoretical physics studies \cite{kerskens2022experimental,neven2024testing,fisher2017we} demonstrate the relationships between quantum theories and the human brain.
Hence, our proposed work represents the first successful machine vision study to realize these hypotheses.

The quantum theory has two \textit{important properties, including superposition and entanglement}.
While the superposition can represent high-dimensional information, the entanglement can compute the connection between entities.
Motivated by quantum theory, we introduce a quantum-inspired network for understanding the vision-brain connection.
The proposed quantum-inspired network includes a new phase-shifting module to calibrate the value of the fMRI voxels, a voxel connection module to compute the connectivity between voxels via entanglement, and a measurement-like projection module to project the information represented in the Hilbert space into the feature space.
It is a simple yet efficient method for computing the connectivity of fMRI voxels in the context of vision-brain understanding.

\noindent
\textbf{Contributions of this Work:}
To contribute to the development of a vision-brain understanding, we introduce a novel Quantum-Brain approach, a quantum-inspired neural network, to learn the connectivity information in brain signals corresponding to human perceptions.
To the best of our knowledge, \textit{the proposed Quantum-Brain model is the first quantum-inspired approach to vision-brain understanding}.
The contributions of this work can be summarized as follows.
First, motivated by the entanglement property in quantum computing, we proposed a novel \textit{Quantum-Inspired Voxel-Controlling Module} for learning the connectivity in the brain signals.
Second, a new \textit{Phase Shifting Module} is introduced to calibrate the voxel values in the brain signals and enhance the robustness of the connectivity extraction.
Third, as the current connectivity information is represented in the quantum space, a new \textit{Measurement-like Projection Module} is presented to transform the extracted information into the feature space.
Finally, through our experiments on the Natural Scene Dataset (NSD) benchmarks \cite{allen2022massive}, including image-brain retrieval and fMRI-to-image reconstruction, our proposed method achieves state-of-the-art performance with the Top-1 accuracy of $95.5\%$ and $95.3\%$ on the image and brain retrieval benchmarks and Inception score of $95.6\%$ on the fMRI-to-image reconstruction benchmark.
Our experimental results demonstrate the effectiveness of brain feature representation for understanding vision and the brain.

\begin{figure*}[t]
    \centering
    \includegraphics[width=0.9\linewidth]{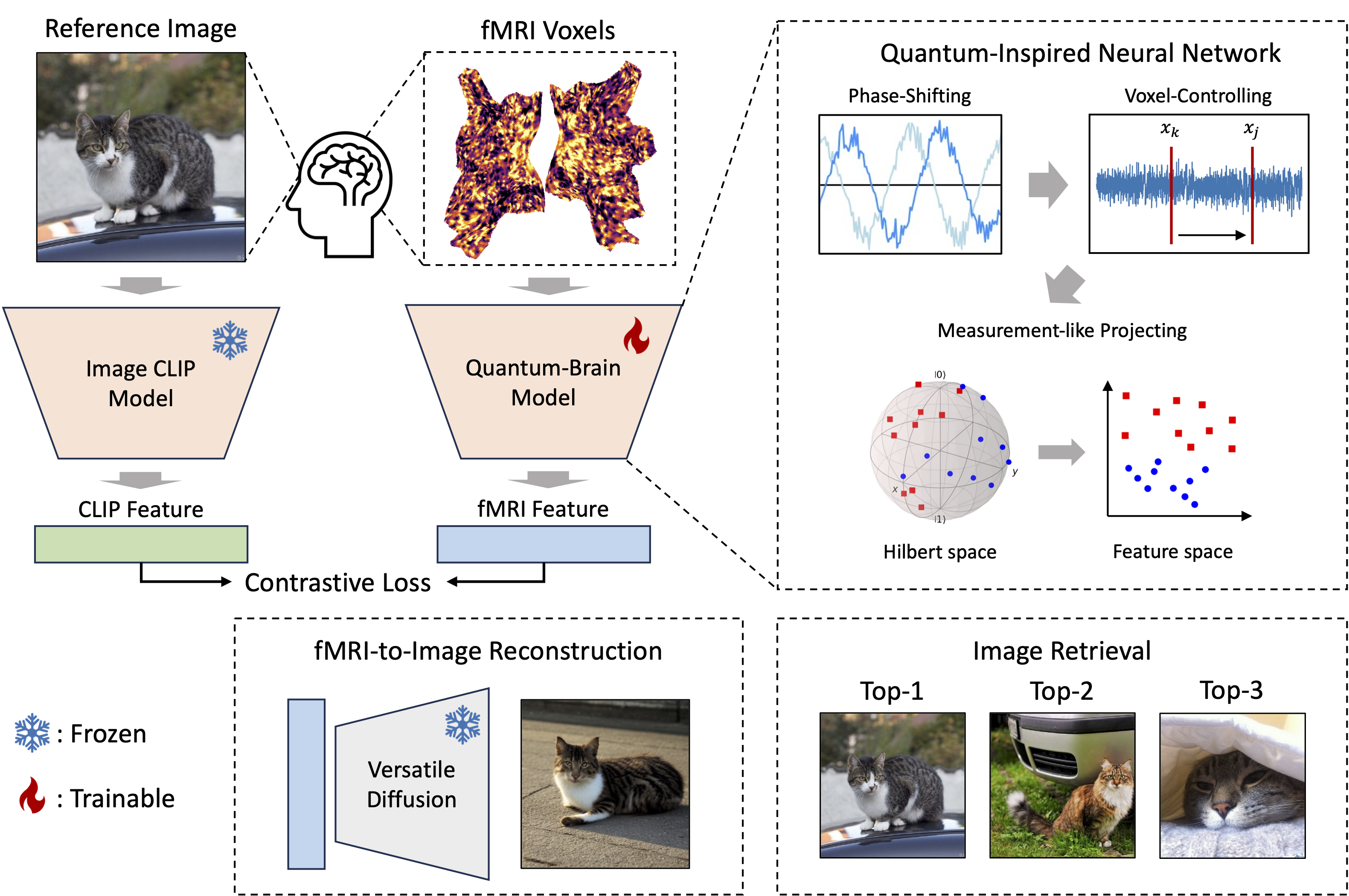}
    \caption{
    \textbf{The overview framework of our proposed quantum-inspired neural network approach to vision-brain understanding.}
    The Phase-Shifting module calibrates the voxel values for better fMRI representation.
    The Voxel-Controlling module computes the connectivities between voxels.
    The Measurement-like Projection module maps the fMRI information from Hilbert space to feature space for semantic feature extraction.
    The extracted semantic features are applied for vision-brain understanding tasks, i.e., fMRI-to-image reconstruction and image-brain retrieval.
    \textbf{Best viewed in color.}
    }
    \label{fig:overview_framework}
% \vspace{-4mm}
\end{figure*}

\section{Related Work}

\subsection{Visual Brain Signal Decoding}
Researchers have shown that visual information, such as spatial position \cite{thirion2006inverse}, orientation \cite{haynes2005predicting,kamitani2005decoding}, and coarse image category \cite{cox2003functional,haxby2001distributed}, can be decoded from fMRI signals using linear models. 
With the advent of generative adversarial networks \cite{goodfellow2020generative}, more advanced decoding methods became possible and enabled researchers to map brain activity to the latent space of these models to reconstruct images such as handwritten digits \cite{schoenmakers2013linear}, human faces \cite{vanrullen2019reconstructing,dado2022hyperrealistic}, and natural scenes \cite{shen2019deep,ozcelik2022reconstruction,seeliger2018generative}. 
Recently, with the development of multimodal contrastive models like CLIP \cite{radford2021learning}, diffusion models \cite{ho2020denoising,song2020improved} like Stable Diffusion \cite{rombach2022high}, and new large-scale fMRI datasets \cite{allen2022massive}, fMRI-to-image reconstructions have achieved an unprecedented level of quality \cite{takagi2023high,ozcelik2023natural,gu2022decoding}.

Lin et al. \cite{lin2022mind} reconstructed seen images by aligning fMRI voxel data into CLIP space and generating outputs via a fine-tuned Lafite GAN \cite{zhou2022towards}.
Moreover, the alignment between the fMRI voxels and the CLIP space enabled the model to perform image-brain retrieval tasks.
Ozcelik and VanRullen \cite{ozcelik2023natural} employed a two-stage approach, utilizing Versatile Diffusion \cite{xu2023versatile} for both low- and high-level processing.
Gu et al. \cite{gu2022decoding} also applied a similar pipeline and extended on Ozcelik et al. \cite{ozcelik2022reconstruction} by implementing IC-GAN \cite{casanova2021instance} for reconstruction. Unlike prior methods, they did not flatten voxels but aligned them to SwAV \cite{caron2020unsupervised} features using surface-based convolutional networks.
Meanwhile, Takagi and Nishimoto \cite{takagi2023high} used ridge regression to align the fMRI voxels into Stable Diffusion latent and CLIP text patients by selecting different voxel regions for various components. 
MinD-Vis \cite{chen2023seeing} addressed fMRI-to-image reconstruction by pre-training a masked brain encoder on a distinct large-scale fMRI dataset rather than the Natural Scenes Dataset \cite{allen2022massive}, producing a more informative latent input for their image reconstruction model.
Building on this, MinD-Video \cite{chen2024cinematic} expanded the MinD-Vis approach to reconstruct video instead of single images.
MindEye \cite{scotti2024reconstructing} enhanced fMRI-to-image reconstruction and image-brain retrieval by separating the feature representation via contrastive learning and diffusion priors.
Additionally, Psychometry \cite{quan2024psychometry}, MindBridge \cite{wang2024mindbridge}, and MindEye2 \cite{scotti2024mindeye2} focused on a unified fMRI feature extraction from multiple subjects.
Meanwhile, previous work analyzed the correlation between fMRI signals and deep learning models \cite{yang2024brain,nguyen2025brainformer,nguyen2024bractive}.
Additionally, with the recent advancements in continual learning methods \cite{truong2025falcon,truong2023fairness} and domain generalization \cite{truong2024ed} across multiple modalities, COBRA \cite{nguyen2024cobra} has studied vision-brain understanding in continual learning settings.
Furthermore, a preliminary study of the human brain in quantum machine learning was introduced \cite{nguyen2024hierarchical}, showing the potential of the relationship between the human brain and quantum theory.

\subsection{Quantum-Inspired Machine Learning}

Recently, quantum computing has been applied to machine learning tasks, i.e., clustering \cite{lloyd2013quantum,nguyen2023quantum,nguyen2024qclusformer,nguyen2024quantum}, principal component analysis \cite{lloyd2014quantum}, least-squares fitting \cite{schuld2016prediction,kerenidis2020quantum}, binary classification \cite{rebentrost2014quantum}, and quadratic optimization \cite{farhi2014quantum,henze2025solving,holliday2025quadro}.
With the rise of quantum neural networks, variational quantum algorithms \cite{panella2011neural,mitarai2018quantum} were introduced to replace the classical deep learning algorithms, including convolutional neural networks \cite{cong2019quantum}, recurrent neural networks \cite{bausch2020recurrent}, generative models \cite{romero2017quantum,huang2021experimental,zhang2024generative,nguyen2025diffusion}, reinforcement learning \cite{chen2020variational,lin2025quantum}, and mixture of experts \cite{tognini2025solving,nguyen2025qmoe}.

Meanwhile, quantum-inspired computing refers to practical methods derived from concepts in quantum computing.
The mathematical formalism of quantum computing has been adopted in various machine learning tasks in classical systems, including natural language processing \cite{li2019cnm}, computer vision \cite{shi2022quantum,tang2022image,zhang2023quantum}, information retrieval \cite{uprety2020survey,van2004geometry}, and multimodal analysis \cite{gkoumas2021quantum,li2021quantum}. 
Van et al. \cite{van2004geometry} applied the quantum probability of object representation in information retrieval tasks. 
Li et al. \cite{li2019cnm} proposed a quantum-inspired network for language models, aiming to achieve better outcome quality and interpretability through semantic matching. 
Li et al. \cite{li2021quantum} investigated multimodal feature fusion methods via a quantum-inspired neural network for conversation emotion recognition. 
Meanwhile, Gkoumas et al. \cite{gkoumas2021quantum} introduced a video semantic recognition network by fusing multimodal information at the decision level. 
Moreover, quantum-inspired methods have also been developed for image-processing tasks. 
Shi et al. \cite{shi2022quantum} presented complex-valued convolutional neural networks to process high-dimensional data and represent a better nonlinear space.
Notably, Tang et al. \cite{tang2022image} utilized a wave-like function to represent an image through phase-shifting operations, thereby learning the relationship between patches.
Meanwhile, Zhang et al. \cite{zhang2023quantum} utilize the quantum state representation to enhance the extraction of hyperspectral image features.

\begin{figure*}[t]
    \centering
    \includegraphics[width=0.9\linewidth]{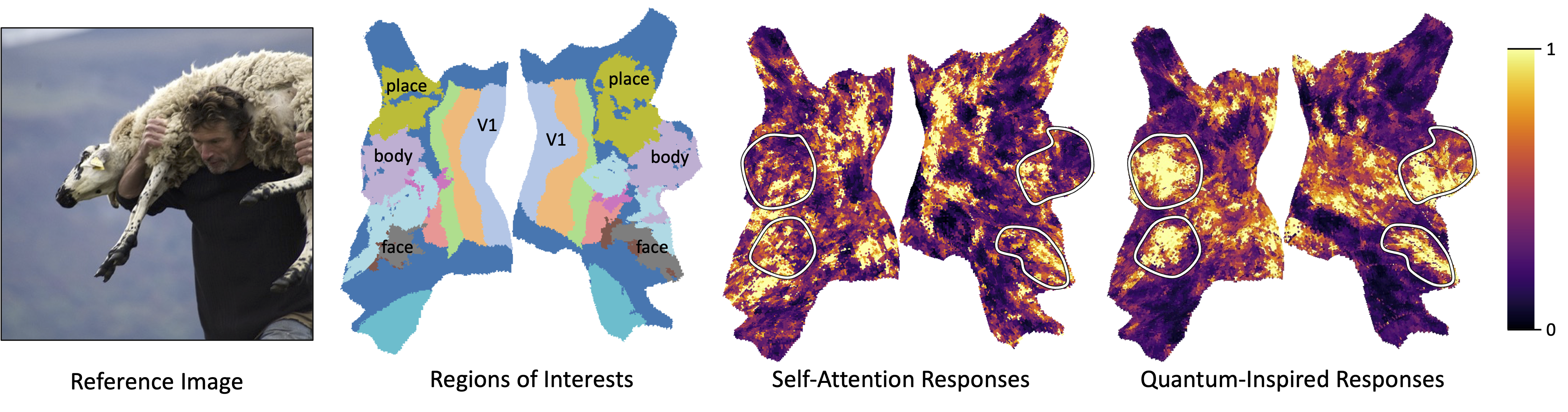}
    % \vspace{-2mm}
    \caption{
    The connectivity map from the V1 brain region, i.e., the low-level perception region, to other regions.
    While the self-attention method fails to illustrate the connectivities between correct specific regions, the proposed quantum-inspired network focuses on the fMRI regions corresponding to the reference image, i.e., the body and face regions.
    \textbf{Best viewed in color.} 
    }
    \label{fig:attention_map_comparison}
% \vspace{-6mm}
\end{figure*}

\section{Preliminaries}

Most quantum computers are constructed based on the quantum computing model \cite{arute2019quantum,nielsen2010quantum}.
A quantum computation includes state initialization, unitary evolution, and measurement. 
The quantum device is first set to the initial state. 
The quantum state is modified by applying unitary operators to quantum devices. 
Then, the initial state of the system is transformed into the final state, which contains the processing results in the quantum device. 
To obtain classical information from the quantum device, measurements are utilized to extract processing results from the final state.

Mathematically, a quantum state can be represented by a state vector in an $N$-dimensional Hilbert space $\mathcal{H}^N$.
We adopt the Dirac notations in quantum theory to denote the quantum states \cite{nielsen2010quantum}.
The notations $|\cdot\rangle$ and $\langle\cdot|$ denote a \textit{Ket} to represent a complex-valued unit vector and a \textit{Bra} to indicate its conjugate transpose.
In detail, given a set of basis state vectors $\{|\phi_n\rangle\}^N_{n=1}$, a quantum state $|\psi\rangle$ can be represented as a combination of basis state vectors as shown in Eqn. \eqref{eq:superposition}.
\begin{equation}
% \small
    |\psi\rangle = \sum^N_{n=1} A_n e^{i \theta_n} |\phi_n\rangle
\label{eq:superposition}
\end{equation}
where $A_n$ is a probability amplitude satisfying $\sum_{n=1}^N |A_n|^2 = 1$, $i$ is the imaginary unit, and $\theta_n$ indicates the phase.
A simple example of basis state vectors for $\mathcal{H}^N$ is the standard unit basis.
In particular, the standard basis of $\mathcal{H}^2$ is formed by column vectors $|0\rangle_2 = (1, 0)^\top$ and $|1\rangle_2 = (0, 1)^\top$.
Here, we use the notation $|n\rangle_N$ to represent the standard basis of $\mathcal{H}^N$.
Respectively, $\langle n|_N$ denotes the conjugate transpose of $|n\rangle_N$.
All components of $|n\rangle_N$ are $0$ except the $n$-th component is $1$.
In quantum theory, Eqn. \eqref{eq:superposition} represents the superposition of basis states.

A unitary operator $U$ denotes a quantum state evolution that transforms the initial states $|\psi\rangle$ to final states $|\psi^\prime\rangle$.
These operators manipulate the phases $\theta_n$ and amplitudes $A_n$ by applying specific physical operations to the quantum system.
In quantum computing, a projection-valued measure (PVM) is a function that can collapse the system state from a superposition of multiple basis states to a single basis state.
The mathematical formulation of PVM is a set of projection operators $\{\Pi_m = |\phi_m\rangle \langle\phi_m|\}_{m=1}^M$.
As described in the \textit{Born rule} \cite{nielsen2010quantum}, $\Pi_m$ projects the quantum state $|\psi^\prime\rangle$ to the basis state $|\phi_m\rangle$ with a probability $P(|\phi_m\rangle)$ as in Eqn. \eqref{eq:pvm}.
\begin{equation}
% \small
    P(|\phi_m\rangle)
    = \langle\phi_m|\psi^\prime\rangle \langle\psi^\prime|\phi_m\rangle
\label{eq:pvm}
\end{equation}

\section{The Proposed Quantum-Brain Approach}

This section presents the vision-brain understanding problem and the limitations of prior vision-brain understanding methods.
Then, we introduce the quantum-inspired neural network based on entanglement.
Finally, we summarize the learning framework for vision-brain understanding.
Fig. \ref{fig:overview_framework} illustrates an overview of the framework for our proposed quantum-inspired neural network approach to vision-brain understanding.

\subsection{Input Modeling}

Let $X_\text{image} \in \mathbb{R}^{H \times W \times 3}$ and $\mathbf{x} \in \mathbb{R}^{C}$ be an image and its corresponding fMRI voxels where $H$, $W$, and $C$ is the height and width of the image and the number of voxels in the fMRI.
An fMRI voxel encoder is applied to extract the features $\mathbf{p} \in \mathbb{R}^D$ of $\textbf{x}$ and learn to align the extracted features to the high-level features $\mathbf{t} \in \mathbb{R}^D$ of $X_\text{image}$ for the semantic understanding, where $D$ is the dimension of the semantic features.
In this work, we utilize CLIP ViT-L/14 \cite{radford2021learning} to construct the semantic feature space of the images, as the CLIP image encoders are trained to maximize the similarity with text captions of the images.

\subsection{Connectivity of Brain Voxels via Entanglement}

\noindent
\textbf{Limitations of Self-Attention. } 
In the Vision-Brain Understanding problem, self-attention in Transformers \cite{vaswani2017attention} is a conventional approach to computing correlations or connectivities between elements.
In the scenario of fMRI voxels, the connectivities between brain voxels depend on the \textit{information processing function} of each brain region.
Each brain region is responsible for processing specific information and can connect with other regions.
For example, a low-level information region that directly processes information from human perception can connect to specific high-level information regions to process more complex semantic information.
A conventional self-attention model learns to find patterns of brain activities without realizing that the connectivities depend on specific positions.
Hence, the self-attention model fails to learn the connectivity in the brain signals.
Fig. \ref{fig:attention_map_comparison} illustrates the comparison between the connectivity learnability of the self-attention method and the proposed quantum-inspired network.

\begin{figure}[t]
% \vspace{-4mm}
\centering
\includegraphics[width=0.8\linewidth]{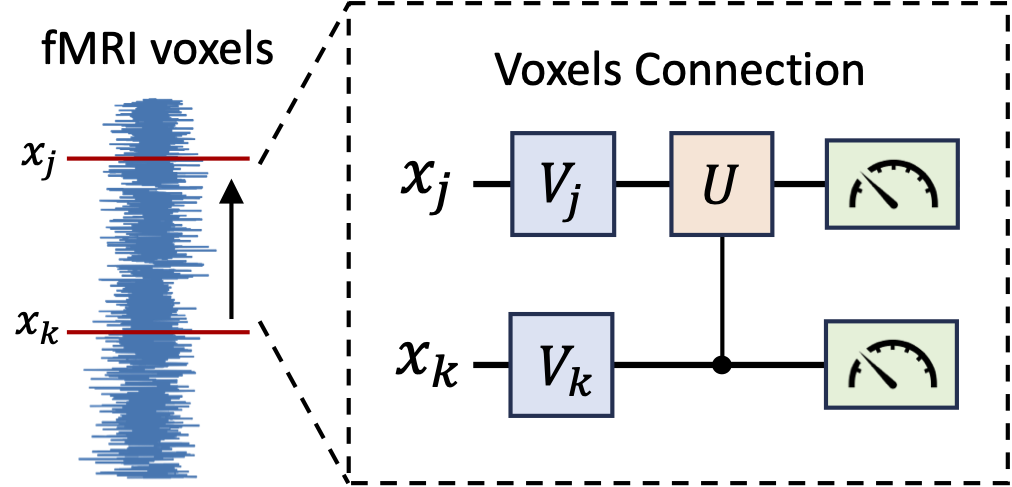}
\caption{
    \textbf{The voxels connection via entanglement.}
    Given two voxels $x_j$ and $x_k$, the connectivity of the voxel $x_k$ to the voxel $x_j$ is computed via the quantum entanglement.
}
\label{fig:voxel_connection}
% \vspace{-6mm}
\end{figure}

\noindent
\textbf{The Connectivity of Entanglement. }
Fig. \ref{fig:voxel_connection} illustrates the connectivity between two voxels via entanglement.
Given two fMRI voxels $x_j$ and $x_k$, we compute the connectivity of the voxel $x_k$ to the voxel $x_j$.
Since the voxels $x_j$ and $x_k$ are raw values representing the brain signal, the phase-shifting operators $V_j$ and $V_k$ are applied to calibrate the voxels into suitable values for the entanglement.
Then, a controlling operator computes the connectivity of the voxel $x_k$ to the voxel $x_j$.
Finally, measurement operators are presented to project the connectivity information into the feature space.

\subsection{Quantum-Inspired Voxel-Controlling Module}

It is crucial to consider the connectivity in the fMRI signal to explore semantic information. The controlling operators in quantum computing can utilize the connectivity between fMRI voxels.
In detail, for each fMRI voxel $x_j$, its value can be formulated into a quantum state $|\psi_j\rangle = \sqrt{1 - x_j} |0\rangle + \sqrt{x_j} |1\rangle$.
Then, the voxel $x_j$ can be controlled by the other voxel $x_k$ via a controlling operator $U_\text{control}^{(j,k)}$ defined as Eqn. \eqref{eq:control_matrix}.
\begin{equation}
% \small
    U_\text{control}^{(j,k)} = 
    \begin{pmatrix}
    I & 0 \\
    0 & U^{(j,k)}
    \end{pmatrix}
\label{eq:control_matrix}
\end{equation}
where $I$ and $U^{(j,k)}$ are $2 \times 2$ identity operator and $2 \times 2$ learnable operator, respectively.
The quantum state representing the connection from the fMRI voxel $x_k$ to the fMRI voxel $x_j$ is computed as Eqn. \eqref{eq:control_function}.
\begin{equation}
% \small
    |\psi_{j,k}\rangle = U_\text{control}^{(j,k)} \left( |\psi_k\rangle \otimes |\psi_j\rangle \right)
\label{eq:control_function}
\end{equation}
where $\otimes$ is the tensor product. 
The higher the voxel value $x_k$ is, the more likely the voxel $x_j$ is affected by $x_k$.
In particular, the operator $U^{(j,k)}$ is defined as controlling weight, i.e., the connectivity weight from the voxel $x_k$ to $x_j$.

Moreover, to enhance the connectivity representation, we apply a Phase-Shifting module, calibrating the voxel values.
In detail, the quantum state $|\psi_j\rangle$ representing the fMRI voxel $x_j$ can be formulated as Eqn. \eqref{eq:phase_shift}.
\begin{equation}
% \small
    |\psi_j\rangle = 
    \sqrt{1 - x_j} e^{i\theta_0^{(j)}} |0\rangle + 
    \sqrt{x_j} e^{i\theta_1^{(j)}} |1\rangle
\label{eq:phase_shift}
\end{equation}
where $\theta_0^{(j)}$ and $\theta_1^{(j)}$ are learnable phase parameters.
This representation can exploit the complex-valued space for connectivity extraction.

\subsection{Measurement-like Projection Module}

The connectivity information represented by the quantum state $|\psi_{j,k}\rangle$ needs to be projected from Hilbert space into the feature space for the later extraction layers.
Inspired by PVM, a measurement-like projection module $f_\text{MPM}$ is introduced to obtain the features of the connectivities between fMRI voxels.
From Eqn. \eqref{eq:pvm}, given a computed state $|\psi_{j,k}\rangle$ and a learnable basis state $|\phi_{j,k}\rangle$, the mathematical formulation of the measurement can be described as Eqn. \eqref{eq:mpm}.
\begin{equation}
% \small
    f_\text{MPM}(|\psi_{j,k}\rangle, |\phi_{j,k}\rangle) = 
    \langle\psi_{j,k}|\phi_{j,k}\rangle 
    \langle\phi_{j,k}|\psi_{j,k}\rangle
\label{eq:mpm}
\end{equation}
While the quantum state $|\psi_{j,k}\rangle$ is represented in the 4-dimensional Hilbert space $\mathcal{H}^4$, the connectivity information can be reduced to a 2-dimensional Hilbert space $\mathcal{H}^2$.
This is because the value of $\sqrt{1 - x_j}$ in $|\psi_k\rangle \otimes |\psi_j\rangle$ is redundant for the projection since the value of $\sqrt{x_j}$ is also in that state. 
In detail, the learnable basis state $|\phi_{j,k}\rangle$ can be defined as
    $|\phi_{j,k}\rangle = 
    \begin{pmatrix}
        0, 1, 0, W_{j,k}
    \end{pmatrix}^\top$.
Then, derived from Eqn. \eqref{eq:mpm}, the connectivity function from $x_k$ to $x_j$ can be formulated as Eqn. \eqref{eq:connectivity_function_j_k}.
\begin{equation}
% \small
\begin{split}
    f(x_j, x_k) = 
    x_j &+ x_j x_k (W_{j,k}^2 - 1) \\
    &+ x_j a_k \cdot 2 W_{j,k} \cos{(\theta_0^{(k)}-\theta_1^{(k)})}
\label{eq:connectivity_function_j_k}
\end{split}
\end{equation}
where $a_k = \sqrt{(1-x_k) x_k}$.
% The full expansion of Eqn. \eqref{eq:connectivity_function_j_k} is described in the Appendix \ref{sec:appendix_expansion}.
When the voxel $x_j$ is highly activated and $a_k$ has a high value, i.e., the voxel $x_k$ is at a moderate activation, the phase information will be applied to compute the connectivity.
Then, the connectivity function can be aggregated from other voxels to the voxel $x_j$ as follows,
\begin{equation}
% \small
\begin{split}
    f(x_j) 
    &= \sum_{k=1}^C c_{j,k} f(x_j, x_k) \\
    &= x_j + x_j \sum_{k=1}^C x_k W^\prime_{j,k} \\
    &\quad\quad+ x_j \sum_{k=1}^C a_k \cos{(\theta_0^{(k)}-\theta_1^{(k)})} W^{\prime\prime}_{j,k}
\end{split}
\end{equation}
where $\sum_{k=1}^C c_{j,k}=1$, $W^\prime_{j,k} = c_{j,k} (W_{j,k}^2 - 1)$, and $W^{\prime\prime}_{j,k} = 2 c_{j,k} W_{j,k}$ are learnable parameters.
Hence, the overall connectivity function of the fMRI voxels is defined as Eqn. \eqref{eq:overall_connectivity_function}.
\begin{equation}
% \small
\begin{split}
    f(\mathbf{x}) &= 
    \mathbf{x} 
    + \mathbf{x} \odot \left( \mathbf{W}^\prime \mathbf{x} \right) \\
    &\quad\quad+ \mathbf{x} \odot \left( \mathbf{W}^{\prime\prime} (\mathbf{a} \odot \cos{(\boldsymbol{\theta}_0 - \boldsymbol{\theta}_1)}) \right)
\label{eq:overall_connectivity_function}
\end{split}
\end{equation}
where $\odot$ is the element-wise product, and $\mathbf{W}^\prime$ and $\mathbf{W}^{\prime\prime}$ are learnable matrices.

\begin{figure}[t]
    \centering
    \includegraphics[width=\linewidth]{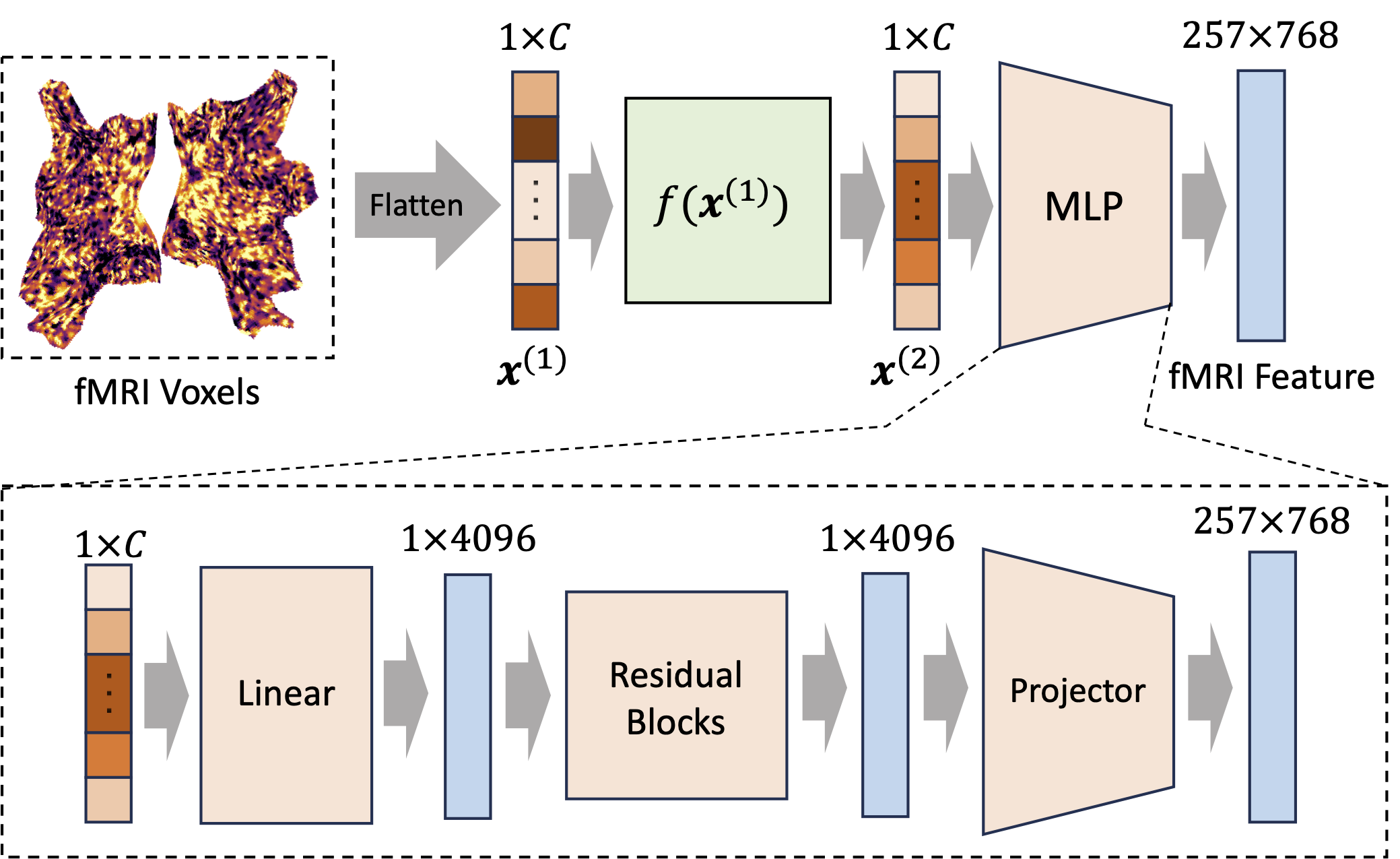}
    \caption{The diagram of the fMRI feature extraction.}
    \label{fig:computing_flow}
\end{figure}

\begin{figure*}[t]
    \centering
    \includegraphics[width=\linewidth]{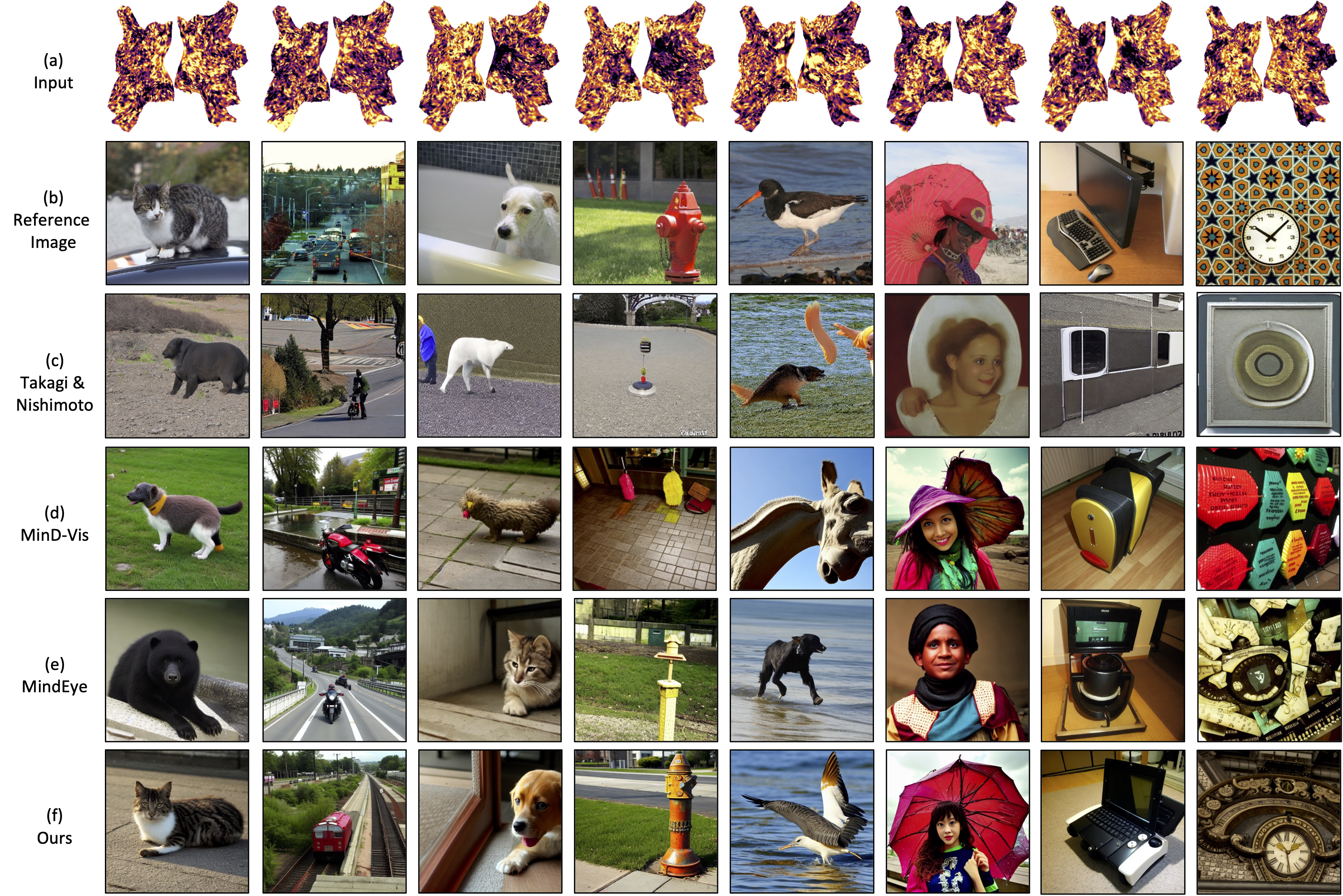}
    % \vspace{-2mm}
    \caption{
    The fMRI-to-image comparison between Takagi \& Nishimoto \cite{takagi2023high}, MinD-Vis \cite{chen2023seeing}, MindEye \cite{scotti2024reconstructing}, and our proposed network on the Subject 1 NSD dataset \cite{allen2022massive}.
    Compared to prior methods, the images reconstructed by our proposed method exhibit more consistent structures and a closer semantic resemblance to the reference images. 
    }
    \label{fig:reconstruction_comparison}
% \vspace{-4mm}
\end{figure*}

\section{fMRI Feature Extraction}

% This section discusses the computing flow of the fMRI feature extraction.
As shown in Figure \ref{fig:computing_flow}, the fMRI voxels are flattened into a voxel vector $\mathbf{x}^{(1)} \in \mathbb{R}^C$ where $C$ is the number of voxels.
The number of voxels varies between different subjects from approximately 13,000 to 16,000 voxels.
The voxel vector $\mathbf{x}^{(1)}$ is computed via connectivity functions $f$ defined as Eqn. \eqref{eq:overall_connectivity_function}.
The computed voxel vector $\mathbf{x}^{(2)}$ is then extracted via a linear layer followed by two residual blocks with a hidden dimension of $4096$.
The feature result of the final block is projected into a semantic feature $\mathbf{p} \in \mathbb{R}^D$.
The feature dimension is denoted as $D = 257 \times 768$ to match the input feature of the Versatile Diffusion model.

\subsection{High-Level Feature Contrastive Learning}

For vision-brain understanding tasks, the features of the fMRI signals should be aligned with the high-level features of the corresponding images.
Contrastive learning is an effective learning method that represents multiple modalities. 
It maximizes the cosine similarity for positive pairs while minimizing the similarity for negative pairs.
The Contrastive Language-Image Pre-trained Model (CLIP) \cite{radford2021learning} is a multimodal contrastive model that aligns the features of images and their captions in texts into the same feature space.
Due to the same feature space mapping, the features extracted from images by CLIP contain high semantic information corresponding to the text captions.

Following the CLIP learning procedure, we use the CLIP loss \cite{radford2021learning} to learn the high-level features of the fMRI signals.
Given $N$ pairs of fMRI feature $\mathbf{p}_i$ and its corresponding image feature $\mathbf{t}_i$, the CLIP loss is defined as Eqn. \eqref{eq:clip_loss}.
\begin{equation}
% \small
\begin{split}
    \mathcal{L}_\text{CLIP} &= 
    \sum_{i=1}^N \left[  
    \log \left(
    \frac{\exp \left( {\mathbf{p}_i^\top \cdot \mathbf{t}_i}/{\tau} \right)}
    {\sum_{j=1}^N \exp \left( {\mathbf{p}_i^\top \cdot \mathbf{t}_j}/{\tau} \right)}
    \right)
    \right] \\
    &\quad+ \sum_{i=1}^N \left[  
    \log \left(
    \frac{\exp \left( {\mathbf{t}_i^\top \cdot \mathbf{p}_i}/{\tau} \right)}
    {\sum_{j=1}^N \exp \left( {\mathbf{t}_i^\top \cdot \mathbf{p}_j}/{\tau} \right)}
    \right)
    \right]
\end{split}
\label{eq:clip_loss}
\end{equation}
where $\tau$ is a temperature hyperparameter.

\section{Experiments}

\subsection{Implementation Details}

Our experiments use CLIP ViT-L/14 \cite{radford2021learning} as a semantic image feature extraction.
The reference images are resized to a resolution of $224 \times 224$.
The quantum-inspired network includes a voxel connection module as shown in Eqn. \eqref{eq:overall_connectivity_function} and a multi-layer perceptron.
%
% The computational details are described in the Appendix \ref{sec:appendix_computing_flow}.
%
The fMRI voxels are extracted into $257\times768$ dimensional features, aligning to the $257\times768$ dimensional hidden CLIP features similar to \cite{ozcelik2023natural,scotti2024reconstructing,wang2024mindbridge}.
We use the CLIP loss with a temperature $\tau = 4 \times 10^{-3}$ for the vision-brain alignment and the pre-trained Versatile Diffusion model \cite{xu2023versatile} for image reconstruction.
For better fMRI-to-image reconstruction, we apply a diffusion prior \cite{ramesh2022hierarchical} to align the output features with the input space of the pre-trained diffusion model.
The learning rate is set to $3\times10^{-4}$ with the Cosine learning rate scheduler \cite{loshchilov2016sgdr}.
The model is optimized using AdamW \cite{loshchilov2017decoupled} with 240 epochs and a batch size of 32.
The proposed model and experiments are implemented and performed in PyTorch \cite{paszke2019pytorch} on a single NVIDIA A100 GPU.

\input{tables/image_reconstruction}

\subsection{Datasets and Benchmarks}

We use the Natural Scenes Dataset (NSD) \cite{allen2022massive} for evaluation.
This public fMRI dataset comprises the brain responses of 8 participants viewing natural images from the MS-COCO dataset \cite{lin2014microsoft}.
We can study the semantic features of brain activities through natural images corresponding to the fMRI signals.
Following \cite{scotti2024reconstructing,ozcelik2023natural,takagi2023high}, we train subject-specific models for each of the four participants separately.
For each participant, there are $25,962$ samples split into $24,980$ training samples and $982$ testing samples, similar to \cite{scotti2024reconstructing,ozcelik2023natural,takagi2023high}.

To evaluate the performance of the proposed quantum-inspired neural network, we perform two vision-brain understanding tasks, including fMRI-to-image reconstruction and image-brain retrieval.
The evaluation protocols and experimental results will be described in the following sections.

\subsection{fMRI-to-Image Reconstruction}

Following MindEye \cite{scotti2024reconstructing}, we use a diffusion prior module to align the fMRI embeddings to the conditional embedding space for the pre-trained image generation.
Similar to MindEye \cite{scotti2024reconstructing}, we use the pre-trained Versatile Diffusion \cite{xu2023versatile} to reconstruct high-quality images from fMRI features.

Fig. \ref{fig:reconstruction_comparison} illustrates the qualitative results of our proposed model in fMRI-to-image reconstruction compared to prior methods \cite{takagi2023high,chen2023seeing,scotti2024reconstructing}.
In addition to the high-quality images, our reconstructed images have a closer semantic meaning to the reference images compared to previous methods.
We also provide quantitative results of the proposed approach as shown in Table \ref{tab:fmri_to_image_reconstruction}.
Following Ozcelik and VanRullen \cite{ozcelik2023natural}, we evaluate the methods with eight different image quality metrics.
PixCorr is a pixel-level correlation metric that measures the similarity between reconstructed images and ground-truth images.
SSIM is the structural similarity index metric.
AlexNet(2) and AlexNet(5) are the 2-way comparisons of the second and fifth layers of AlexNet, respectively.
Inception is a two-way comparison of the last pooling layer in InceptionV3.
CLIP is the 2-way comparison of the output layer of the CLIP-Vision model. 
EffNet-B and SwAV are distance metrics gathered from EfficientNet-B1 and SwAV-ResNet50 models.
The first four are low-level metrics, while the last four express higher-level properties.
As illustrated in Table \ref{tab:fmri_to_image_reconstruction}, our proposed model extracts fMRI information and achieves better high-level similarities to the corresponding images compared to prior methods.
Meanwhile, the proposed model obtains competitive low-level results.
It shows that the features extracted by the proposed quantum-inspired neural network are well-aligned with the semantic feature space of the images.

\subsection{Image-Brain Retrieval}

Image-brain retrieval benchmarks assess the level of fine-grained visual information contained in brain embeddings.
Following Lin et al. \cite{lin2022mind}, the image-brain retrieval benchmarks evaluate the models in two separate procedures.
For the image retrieval task, each testing fMRI sample is extracted into an fMRI embedding. 
Then we compute the cosine similarity to the CLIP image embeddings of its corresponding image and 299 other random images.
For each testing sample, a correct retrieval is determined when the cosine similarity of the fMRI embedding and its corresponding image embedding is the highest.
We average the retrieval results across all testing samples and repeat the whole process 30 times.
For brain retrieval, the procedure is similar to the image retrieval, except we randomly select 299 other fMRI samples.

\input{tables/image_retrieval}
As shown in Table \ref{tab:image_brain_retrieval}, our model outperforms previous methods \cite{lin2022mind,ozcelik2023natural,scotti2024reconstructing} by a large margin.
On the image-brain retrieval benchmark, the proposed approach achieves higher top-1 accuracy of $95.5\%$ and $95.3\%$ on the image and brain retrieval evaluations, respectively.
The results show that the proposed model exhibits better and more balanced performance on the image-brain retrieval benchmark.

\subsection{Ablation Studies}

\input{tables/ablation_studies}
Our ablation experiments study the effectiveness of our proposed modules on the NSD image-brain retrieval benchmark, as shown in Table \ref{tab:ablation_studies}.

\noindent
\textbf{Effectiveness of Phase-Shifting Module. }
As shown in Table \ref{tab:ablation_studies}, the Phase-Shifting module calibrates and better represents the brain voxel values for later information extraction.
In detail, the top-1 accuracies of image retrieval have been increased by $0.9\%$, $1.7\%$, and $1.5\%$ for the three settings.
Moreover, the Phase-Shifting module enables the Voxel-Controlling module to extract brain information more effectively, resulting in an increase in the top-1 accuracy of image and brain retrieval from $0.7\%$ to $1.5\%$ and from $0.7\%$ to $4.2\%$.

\noindent
\textbf{Effectiveness of Controlling Module. }
Table \ref{tab:ablation_studies} illustrates the impact of the Voxel-Controlling module.
As shown in our results, the Voxel-Controlling module has improved the top-1 accuracies of image and brain retrieval from $90.7\%$ and $86.4\%$ to $91.4\%$ and $87.1\%$, respectively, without the Phase-Shifting module, and from $91.6\%$ and $87.2\%$ to $93.1\%$ and $91.4\%$ with the Phase-Shifting module.

\noindent
\textbf{Effectiveness of Measurement-like Projection Module. }
As reported in Table \ref{tab:ablation_studies}, the Measurement-like Projection Module helps to transform the information in the Hilbert space into the feature space.
Hence, the extracted information can be well-aligned with the semantic information obtained from the image CLIP features.
In particular, the top-1 accuracies of image and brain retrieval have increased by $2.6\%$ and $1.5\%$ without the Phase-Shifting module, and by $2.4\%$ and $3.9\%$ with the Phase-Shifting module.

% \begin{figure*}[t]
%     \centering
%     \includegraphics[width=1.0\linewidth]{figures/fmri_to_images_reconstruction_subjects.png}
%     \caption{The NSD fMRI-to-image reconstruction examples of the four subjects.}
%     \label{fig:reconstruction_subjects}
% \end{figure*}

% \input{tables/subject_specific}

% \subsection{Subject-Specific Results}

\section{Conclusions}

This work has introduced Quantum-Brain, a novel quantum-inspired neural network for vision-brain understanding. Motivated by the entanglement properties in quantum computing theory, we have proposed a novel Quantum-Inspired Voxel-Controlling module to compute the connectivities between fMRI voxels represented in the Hilbert space.
Then, a new Phase-Shifting module has been introduced to enable the Quantum-Inspired Voxel-Controlling module to learn connectivities more effectively. Finally, we have presented a novel Measurement-like Projection Module to transform the connectivity information from the Hilbert space to the feature space. Our experimental results have demonstrated the effectiveness and significance of our proposed quantum-inspired neural network.

\noindent
\textbf{Limitations: }
This study used a specific network design and hyperparameters to support our hypothesis.
Our experiments are limited to the standard scale of the benchmarks due to computational limitations.
However, our proposed approach can be generalized for larger-scale benchmarks based on our theoretical analysis.

\noindent
\textbf{Broader Impacts: }
Our contributions open a novel paradigm for modeling complex brain connectivities, leveraging principles such as entanglement and superposition to enhance brain signal interpretation, and revolutionize the development of brain-computer interfaces by bridging theoretical quantum models with practical neuroscience applications.

{
    \small
    \bibliographystyle{IEEEtran}
    \bibliography{IEEE_main}
}

\begin{IEEEbiography}
[{\includegraphics[width=1in,height=1.25in,clip,keepaspectratio]{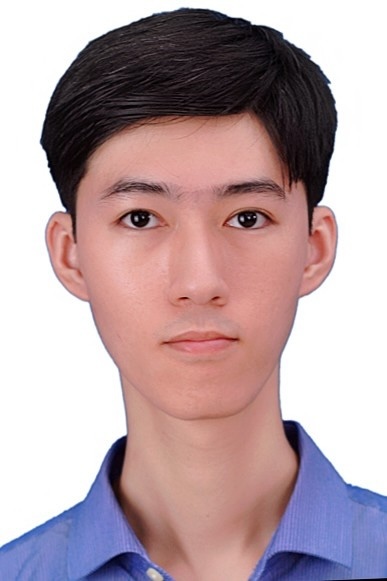}}]
{Hoang-Quan Nguyen} is a PhD candidate at the University of Arkansas, where he is supervised by Dr. Khoa Luu. He is a research assistant at the Computer Vision and Image Understanding Lab. In 2022, he received his B.Sc. degree in Computer Science from the Honors Program at the University of Science, VNU-HCM, under the supervision of Dr. Khoa Luu and Dr. Son Tran. His research interest includes computer vision, deep learning, and quantum machine learning.
\end{IEEEbiography}

\begin{IEEEbiography}
[{\includegraphics[width=1in,height=1.25in,clip,keepaspectratio]{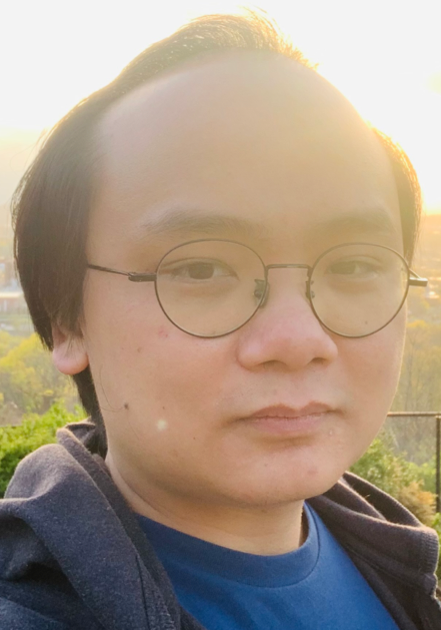}}]
{Xuan-Bac Nguyen} received the B.Sc. degree in electronics and telecommunications from the VNU University of Engineering and Technology in 2015 and the M.Sc. degree in computer science from the Department of Electrical and Computer Engineering, Chonnam National University, South Korea, in 2020. He is currently pursuing the Ph.D. degree with the Department of Electrical Engineering and Computer Science at the University of Arkansas. In 2016, he was a Software Engineer in Yokohama, Japan. His research interests include quantum machine learning, face recognition, facial expression, and medical image processing.
\end{IEEEbiography}

\begin{IEEEbiography}
[{\includegraphics[width=1in,height=1.25in,clip,keepaspectratio]{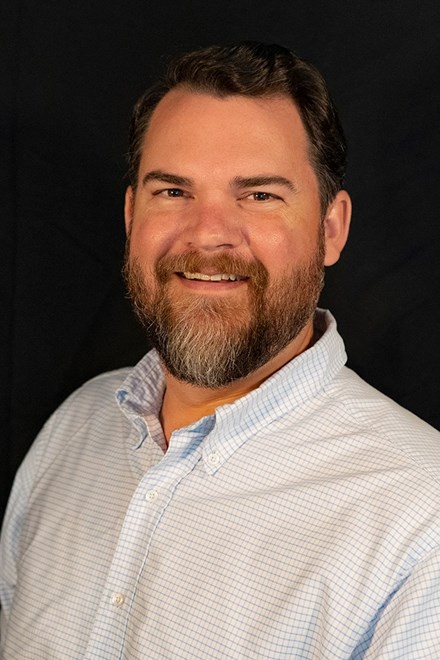}}]
{Hugh Churchill} received the Ph.D. degree in physics from Harvard University in 2012. He was a Pappalardo Postdoctoral Fellow in physics at MIT. He is currently a Professor and the 21st Century Chair of Nanophysics with the Department of Physics, University of Arkansas (UA). He is also the Associate Director of Operations at MonArk NSF Quantum Foundry. Since 2015, he has led the Quantum Device Laboratory, UA. His research interests include quantum materials and devices, low-dimensional materials, quantum transport, optoelectronics, and automation of experiments.
\end{IEEEbiography}

\begin{IEEEbiography}
[{\includegraphics[width=1in,height=1.25in,clip,keepaspectratio]{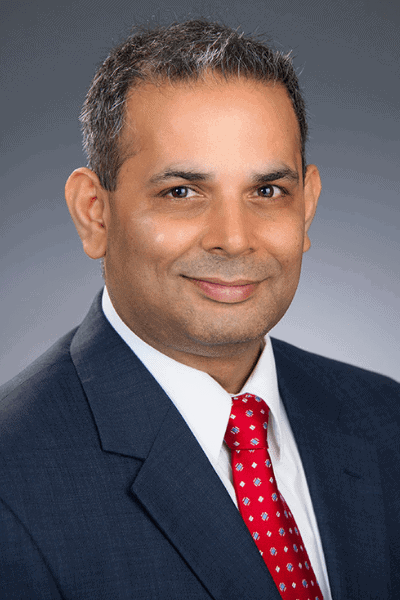}}]
{Arabinda Kumar Choudhary} is Chair of Radiology at SUNY Upstate Medical University and a board-certified specialist in clinical informatics, neuroradiology, and pediatric radiology. He has previously served as Chair of Radiology at UAMS and at Nemours/Alfred I. duPont Hospital for Children. His research focuses on pediatric neuroimaging, imaging informatics, and diagnostic approaches to abusive head trauma. Dr. Choudhary received his medical training at Calcutta University and the University of Cambridge, completed fellowships at Cincinnati Children’s Hospital, and holds an MBA in health care management from the Wharton School.
\end{IEEEbiography}

\begin{IEEEbiography}
[{\includegraphics[width=1in,height=1.25in,clip,keepaspectratio]{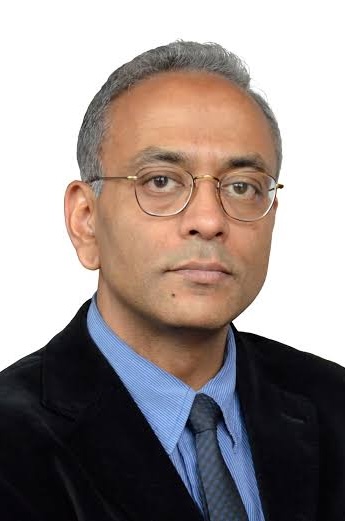}}]
{Pawan Sinha} is a professor of vision and computational neuroscience in the Department of Brain and Cognitive Sciences at MIT. He received his undergraduate degree in computer science from the Indian Institute of Technology, New Delhi and his Masters and doctoral degrees from the Department of Computer Science at MIT. He was at the University of California, Berkeley for the first year of his graduate studies.
Pawan has served on the program committees for prominent scientific conferences on object and face recognition and is currently a member of the editorial board of ACM's Journal of Applied Perception. He is a recipient of the Pisart Vision Award from the Lighthouse Guild International, the PECASE – the highest US Government award for young scientists, the Alfred P. Sloan Foundation Fellowship in Neuroscience, the John Merck Scholars Award for research on developmental disorders, the Jeptha and Emily Wade Award for creative research, the James McDonnell Scholar Award, the Troland Award from the National Academies, the Global Indus Technovator Award and the Distinguished Alumnus Award from IIT Delhi. Pawan's teaching has been recognized by Departmental honors and the Dean’s Award for Advising and Teaching at MIT.
\end{IEEEbiography}

\begin{IEEEbiography}
[{\includegraphics[width=1in,height=1.25in,clip,keepaspectratio]{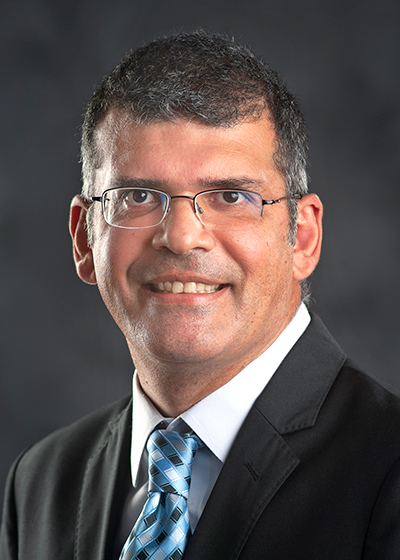}}]
{Samee U. Khan} received a Ph.D. in 2007 from the University of Texas, Arlington, TX. He is Professor and Head of the Mike Wieges Department of Electrical and Computer Engineering at Kansas State University (K-State). Before joining K-State, he was a faculty member at Mississippi State University (MSU), serving as Department Head and the James W. Bagley Chair Professor from 2020 to 2024. He started his career at North Dakota State University in 2008 and rose through the ranks to become the Walter B. Booth Professor. While at NDSU, he was assigned to the National Science Foundation (2016-2020) as Cluster Lead for Computer Systems Research within the Computer and Network Systems Division. 
His research interests include computer system optimization, robustness, and security. His work has appeared in over 475 publications. He is the associate editor of IEEE Transactions on Cloud Computing and the Journal of Parallel and Distributed Computing. He is a Fellow of the IET and BCS, a Distinguished Member of the ACM, and a Senior Member of the IEEE. 
\end{IEEEbiography}

\begin{IEEEbiography}
[{\includegraphics[width=1in,height=1.25in,clip,keepaspectratio]{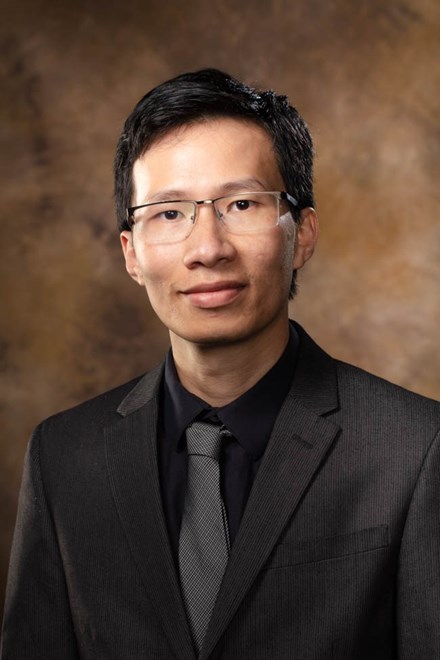}}]
{Khoa Luu} was the Research Project Director at the Cylab Biometrics Center, Carnegie Mellon University (CMU), Pittsburgh, PA, USA. He is currently an Assistant Professor and the Director of the Computer Vision and Image Understanding (CVIU) Laboratory, Department of Electrical Engineering and Computer Science (EECS), University of Arkansas (UA), Fayetteville, AR, USA. He is also with the MonARK NSF Quantum Foundry. He has received eight patents and three best paper awards and co-authored more than 120 papers in conferences, technical reports, and journals. His research interests include quantum machine learning, biometrics, smart health, and precision agriculture. 
He was a PC member of AAAI, ICPRAI, in 2020, 2021, and 2022. He was the Vice-Chair of the Montreal Chapter of the IEEE Systems, Man, and Cybernetics Society in Canada, from September 2009 to March 2011; the Technical Program Chair of the IEEE GreenTech Conference 2024; and a Co-Organizer and the Chair of the CVPR Precognition Workshop, from 2019 to 2024, the MICCAI Workshop, from 2019 to 2020, and ICCV Workshop, in 2021. He is also the Area Chair in CVPR 2023, CVPR 2024, NeurIPS 2024, WACV 2025, and ICLR 2025. He has been an Active Reviewer for several top-tier conferences and journals, such as CVPR, ICCV, ECCV, NeurIPS, ICLR, FG, BTAS, IEEE TRANSACTIONS ON PATTERN ANALYSIS AND MACHINE INTELLIGENCE, IEEE TRANSACTIONS ON IMAGE PROCESSING, IEEE ACCESS, Pattern Recognition, Image and Vision Computing, Journal of Signal Processing, and Journal of Intelligence Review. He is an Associate Editor of IEEE ACCESS and the Multimedia Tools and Applications (Springer Nature).
\end{IEEEbiography}

\input{appendices}

\end{document}

%% file: tables/image_reconstruction.tex
% \usepackage{multirow}
\begin{table*}[t]
\centering
\caption{
Quantitative comparison of fMRI-to-image reconstruction on the NSD benchmark.
The \textbf{bolded} results indicate the best results, and the \underline{underlined} results indicate the secondary best results.
}
\resizebox{\linewidth}{!}{
\begin{tabular}{l|cccc|cccc}
\Xhline{2\arrayrulewidth}
\multirow{2}{*}{\textbf{Method}} & \multicolumn{4}{c}{\textbf{Low-level}} & \multicolumn{4}{c}{\textbf{High-level}} \\
 &
  \multicolumn{1}{c}{\textbf{PixCorr}$\uparrow$} &
  \multicolumn{1}{c}{\textbf{SSIM}$\uparrow$} &
  \multicolumn{1}{c}{\textbf{AlexNet(2)}$\uparrow$} &
  \multicolumn{1}{c}{\textbf{AlexNet(5)}$\uparrow$} &
  \multicolumn{1}{c}{\textbf{InceptionV3}$\uparrow$} &
  \multicolumn{1}{c}{\textbf{CLIP}$\uparrow$} &
  \multicolumn{1}{c}{\textbf{EffNet-B}$\downarrow$} &
  \multicolumn{1}{c}{\textbf{SwAV}$\downarrow$} \\
\hline
Lin et al. \cite{lin2022mind} & - & - & - & - & 78.2\% & - & - & - \\
Takagi \& Nishimoto \cite{takagi2023high}  & - & - & 83.0\% & 83.0\% & 76.0\% & 77.0\% & - & - \\
Gu et al. \cite{gu2022decoding}  & .150 & .325 & - & - & - & - & .862 & .465 \\
Ozcelik \& VanRullen \cite{ozcelik2023natural} & .254 & \textbf{.356} & 94.2\% & 96.2\% & 87.2\% & 91.5\% & .775 & .423 \\
MinD-Vis \cite{chen2023seeing} & .145 & .294 & 82.4\% & 92.4\% & 89.9\% & 90.2\% & .715 & .413 \\
MindEye \cite{scotti2024reconstructing} & \textbf{.309} & .323 & 94.7\% & \underline{97.8\%} & 93.8\% & 94.1\% & .645 & \underline{.367} \\
MindBridge (Single) \cite{wang2024mindbridge} & .148 & .259 & 86.9\% & 95.3\% & 92.2\% & 94.3\% & .713 & .413 \\
DREAM \cite{xia2024dream} & .288 & .338 & \underline{95.0\%} & 97.5\% & \underline{94.8\%} & \underline{95.2\%} & \underline{.638} & .413 \\
\hline
\textbf{Ours}                    & \underline{.308} & \underline{.343} & \textbf{95.3\%} & \textbf{97.9\%} & \textbf{95.6\%} & \textbf{96.7\%} & \textbf{.611} & \textbf{.343} \\
\Xhline{2\arrayrulewidth}
\end{tabular}
}
% \vspace{-6mm}
\label{tab:fmri_to_image_reconstruction}
\end{table*}

%% file: tables/image_retrieval.tex
\begin{table}[t]
% \begin{wraptable}{r}{0.45\linewidth}
% \vspace{-8mm}
\centering
\caption{Image-Brain retrieval results on the NSD benchmark.}
% \vspace{-2mm}
\resizebox{\linewidth}{!}{
\begin{tabular}{lccc}
\Xhline{2\arrayrulewidth}
\textbf{Method} & \textbf{Model} & \textbf{Image}$\uparrow$  & \textbf{Brain}$\uparrow$  \\
\hline
Lin et al. \cite{lin2022mind} & MLP & 11.0\% & 49.0\% \\
Ozcelik... \cite{ozcelik2023natural} & Linear Regression & 21.1\% & 30.3\% \\
MinD-Vis \cite{chen2023seeing} & Transformer & 91.6\% & 85.9\% \\
MindEye \cite{scotti2024reconstructing} & MLP + Projector + Prior & 93.6\% & 90.1\% \\
\hline
\textbf{Ours}           & \textbf{Quantum-Brain} & \textbf{95.5\%} & \textbf{95.3\%} \\
\Xhline{2\arrayrulewidth}
\end{tabular}
}
% \vspace{-4mm}
\label{tab:image_brain_retrieval}
% \end{wraptable}
\end{table}

%% file: tables/ablation_studies.tex
\begin{table}[t]
% \begin{wraptable}{r}{0.45\linewidth}
% \vspace{-8mm}
\centering
\caption{Effectiveness of our method on the NSD Image-Brain Retrieval.}
% \vspace{1mm}
\resizebox{\linewidth}{!}{
\begin{tabular}{c|c|c|cc}
\Xhline{2\arrayrulewidth}
\makecell{\textbf{Phase-} \\ \textbf{Shifting}} & \makecell{\textbf{Voxel-} \\ \textbf{Controlling}} & \makecell{\textbf{Measurement-like}\\\textbf{Projection}} & \textbf{Image}$\uparrow$  & \textbf{Brain}$\uparrow$  \\
\hline
 & & & 90.7\% & 86.4\% \\
 & \checkmark & & 91.4\% & 87.1\% \\
 & \checkmark & \checkmark & 94.0\% & 88.6\% \\
\hline
\checkmark & & & 91.6\% & 87.2\% \\
\checkmark & \checkmark & & 93.1\% & 91.4\% \\
\checkmark & \checkmark & \checkmark & \textbf{95.5\%} & \textbf{95.3\%} \\
\Xhline{2\arrayrulewidth}
\end{tabular}
}
% \vspace{-4mm}
\label{tab:ablation_studies}
% \end{wraptable}
\end{table}

%% file: appendices.tex
\newpage

{\appendices
\section{Expansion Details of Equation \eqref{eq:connectivity_function_j_k}}
\label{sec:appendix_expansion}

From Eqn. \eqref{eq:control_function}, the quantum state $|\psi_{j,k}\rangle$ can be represented as Eqn. \eqref{eq:superposition_representation}.
\begin{equation}
    |\psi_{j,k}\rangle = 
    \begin{pmatrix}
    \sqrt{(1 - x_k)(1 - x_j)} e^{i(\theta^{(k)}_0 + \theta^{(j)}_0)} \\
    \sqrt{(1 - x_k)x_j} e^{i(\theta^{(k)}_0 + \theta^{(j)}_1)} \\
    \sqrt{x_k(1 - x_j)} e^{i(\theta^{(k)}_1 + \theta^{(j)}_0)} \\
    \sqrt{x_k x_j} e^{i(\theta^{(k)}_1 + \theta^{(j)}_1)}
    \end{pmatrix}
\label{eq:superposition_representation}
\end{equation}
Then the inner product between the basis state $|\phi_{j,k}\rangle$ and $|\psi_{j,k}\rangle$ is computed as Eqn. \eqref{eq:inner_product}.
\begin{equation}
\begin{split}
    \langle \phi_{j,k} | \psi_{j,k} \rangle 
    &= \sqrt{(1 - x_k)x_j} e^{i(\theta^{(k)}_0 + \theta^{(j)}_1)} \\
    &\quad\quad + W_{j,k} \sqrt{x_k x_j} e^{i(\theta^{(k)}_1 + \theta^{(j)}_1)}
\end{split}
\label{eq:inner_product}
\end{equation}
Respectively, the inner product $\langle \psi_{j,k} | \phi_{j,k} \rangle$ is computed as follows,
\begin{equation}
\begin{split}
    \langle \psi_{j,k} | \phi_{j,k} \rangle
    &= \langle \phi_{j,k} | \psi_{j,k} \rangle ^* \\
    &= \sqrt{(1 - x_k)x_j} e^{-i(\theta^{(k)}_0 + \theta^{(j)}_1)} \\
    &\quad\quad + W_{j,k} \sqrt{x_k x_j} e^{-i(\theta^{(k)}_1 + \theta^{(j)}_1)}
\end{split}
\end{equation}
where $\alpha^*$ indicates the complex conjugate of $\alpha$.
Next, the connectivity function from $x_k$ to $x_j$ can be formulated as follows,
\begin{equation}
\begin{split}
    f(x_j, x_k) 
    &= f_\text{MPM}(|\psi_{j,k}\rangle, |\phi_{j,k}\rangle) \\
    &= \langle \psi_{j,k} | \phi_{j,k} \rangle \langle \phi_{j,k} | \psi_{j,k} \rangle \\
    &= (1 - x_k) x_j + W_{j,k}^2 x_k x_j \\
    &\quad\quad+ W_{j,k} x_j \sqrt{(1 - x_k) x_k} \\
    &\quad\quad\quad \cdot \left( 
    e^{i(\theta^{(k)}_0 - \theta^{(k)}_1)} + 
    e^{-i(\theta^{(k)}_0 - \theta^{(k)}_1)} 
    \right) \\
    &= x_j + x_j x_k (W_{j,k}^2 - 1) \\
    &\quad+ 2 W_{j,k} x_j \sqrt{(1 - x_k) x_k} \cos{(\theta^{(k)}_0 - \theta^{(k)}_1)}
\end{split}
\end{equation}
since $e^{i \beta} + e^{-i \beta} = 2 \cos{(\beta)}$ from Euler's formula.
Then, we can define $a_k = \sqrt{(1 - x_k) x_k}$ as shown in Eqn. \eqref{eq:connectivity_function_j_k}.

% \section{Quantum-Inspired Responses Computing}
\section{fMRI Responses Computing}

\noindent
\textbf{Quantum-Inspired Responses Computing. }
To compute the quantum-inspired responses from the V1 region to other regions as illustrated in Fig. \ref{fig:intro} and Fig. \ref{fig:attention_map_comparison}, we drop out the values of all voxels except V1 as shown in Eqn. \eqref{eq:drop_out}.
\begin{equation}
    \mathbf{x}^\prime = \mathbf{b}_\text{mask} \odot \mathbf{x}
\label{eq:drop_out}
\end{equation}
where $\mathbf{b}_\text{mask}$ is a masked vector of the V1 region, i.e., $\mathbf{b}_\text{mask}^{(i)} = 1$ if the voxel $x_i$ is in the V1 region and otherwise $\mathbf{b}_\text{mask}^{(i)} = 0$.
Then, we utilize the Eqn. \eqref{eq:overall_connectivity_function} to compute the responses $f(\mathbf{x}^\prime)$ as shown in Eqn. \eqref{eq:v1_responses}.
\begin{equation}
\begin{split}    
    f(\mathbf{x}^\prime) &= 
    \mathbf{x} 
    + \mathbf{x} \odot \left( \mathbf{W}^\prime \mathbf{x}^\prime \right) \\
    &\quad\quad + \mathbf{x} \odot \left( \mathbf{W}^{\prime\prime} (\mathbf{a}^\prime \odot \cos{(\boldsymbol{\theta}_0 - \boldsymbol{\theta}_1)}) \right)
\end{split}
\label{eq:v1_responses}
\end{equation}
where $\mathbf{a}^\prime = \sqrt{(1-\mathbf{x}^\prime) \odot \mathbf{x}^\prime}$.

\noindent
\textbf{Attention Map for Responses Computing. }
To compute the attention map as shown in Fig. \ref{fig:intro} and Fig. \ref{fig:attention_map_comparison}, we compute the attention scores obtained from the last layer of the Transformer model.
Since we want to illustrate the connectivities from region V1 to other regions, we average the attention of the voxels in region V1 only to obtain the attention result.

% \section{Training Procedure}

% \input{code/train}

% \begin{figure}[t]
%     \centering
%     \includegraphics[width=\linewidth]{figures/computing_flow.pdf}
%     \caption{The diagram of the fMRI feature extraction.}
%     \label{fig:computing_flow}
% \end{figure}

\begin{figure*}[t]
    \centering
    \includegraphics[width=1.0\linewidth]{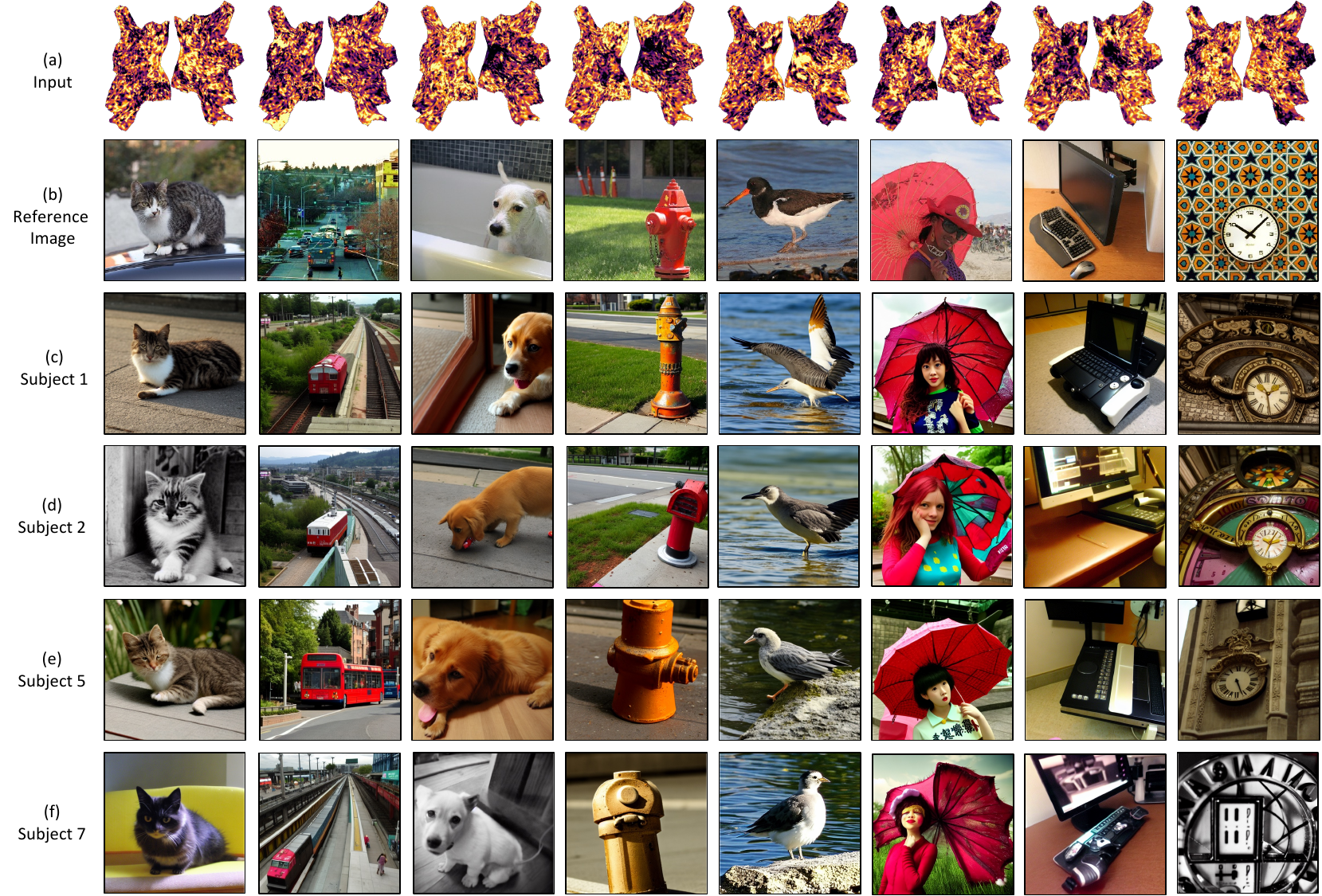}
    \caption{The NSD fMRI-to-image reconstruction examples of the four subjects.}
    \label{fig:reconstruction_subjects}
\end{figure*}

\input{tables/subject_specific}

\section{Regions of Interest}

In the Natural Scene Dataset, we utilize three regions of interest (ROIs), i.e., faces, places, and bodies, to visualize the performance of our model.
These ROIs are obtained via functional localizer experiments (fLoc)
\footnote{
In the NSD dataset, this information can be obtained in {\tt{nsddata/ppdata/subjAA/*/roi/}}, where {\tt{AA}} is the name of the subject.
}.
Note that these regions are not utilized for the training process.

% \section{fMRI Feature Extraction}
% \label{sec:appendix_computing_flow}

% This section discusses the computing flow of the fMRI feature extraction.
% As shown in Figure \ref{fig:computing_flow}, the fMRI voxels are flattened into a voxel vector $\mathbf{x}^{(1)} \in \mathbb{R}^C$ where $C$ is the number of voxels.
% The number of voxels varies between different subjects from approximately 13,000 to 16,000 voxels.
% The voxel vector $\mathbf{x}^{(1)}$ is computed via connectivity functions $f$ defined as Eqn. \eqref{eq:overall_connectivity_function}.
% The computed voxel vector $\mathbf{x}^{(2)}$ is then extracted via a linear layer followed by two residual blocks with a hidden dimension of $4096$.
% The feature result of the final block is projected into a semantic feature $\mathbf{p} \in \mathbb{R}^D$.
% The feature dimension is denoted as $D = 257 \times 768$ to match the input feature of the Versatile Diffusion model.

\section{Subject-Specific Results}

Fig. \ref{fig:reconstruction_subjects} shows the fMRI-to-image reconstruction examples of the four subjects 1, subject 2, subject 5, and subject 7.
Additionally, Table \ref{tab:subject_specific} reports the fMRI-to-image reconstruction and image-brain retrieval results of individual subjects.
}

%% file: tables/subject_specific.tex
% \usepackage{multirow}
\begin{table*}[t]
\centering
\caption{Quantitative results of fMRI-to-image reconstruction on the NSD benchmark with different subjects.
}
\resizebox{\linewidth}{!}{
\begin{tabular}{l|cccc|cccc|cc}
\Xhline{2\arrayrulewidth}
\multirow{2}{*}{\textbf{Subject}} & \multicolumn{4}{c}{\textbf{Low-level}} & \multicolumn{4}{c}{\textbf{High-level}} & \multicolumn{2}{c}{\textbf{Retrieval}} \\
 &
  \multicolumn{1}{c}{\textbf{PixCorr}$\uparrow$} &
  \multicolumn{1}{c}{\textbf{SSIM}$\uparrow$} &
  \multicolumn{1}{c}{\textbf{AlexNet(2)}$\uparrow$} &
  \multicolumn{1}{c}{\textbf{AlexNet(5)}$\uparrow$} &
  \multicolumn{1}{c}{\textbf{InceptionV3}$\uparrow$} &
  \multicolumn{1}{c}{\textbf{CLIP}$\uparrow$} &
  \multicolumn{1}{c}{\textbf{EffNet-B}$\downarrow$} &
  \multicolumn{1}{c}{\textbf{SwAV}$\downarrow$} & 
  \multicolumn{1}{c}{\textbf{Image}$\uparrow$} & 
  \multicolumn{1}{c}{\textbf{Brain}$\uparrow$} \\
\hline
Subj01 & .372 & .361 & 96.9\% & 98.8\% & 96.7\% & 96.7\% & .591 & .331 & 98.6\% & 97.4\% \\
Subj02 & .304 & .356 & 97.1\% & 98.7\% & 95.2\% & 95.8\% & .625 & .337 & 97.7\% & 97.6\% \\
Subj05 & .270 & .331 & 93.2\% & 97.1\% & 96.6\% & 97.6\% & .584	& .333 & 93.9\% & 93.7\% \\
Subj07 & .286 & .323 & 94.1\% & 96.9\% & 93.7\% & 96.5\% & .643 & .371 & 91.6\% & 92.5\% \\
\Xhline{2\arrayrulewidth}
\end{tabular}
}
\label{tab:subject_specific}
\end{table*}